\documentclass{article}
	\usepackage{spconf} 
\usepackage[english]{babel}
\usepackage{ifpdf}
\usepackage{cite} 
\usepackage{url}
\usepackage{hyperref}
\ifpdf
	\usepackage[pdftex]{graphicx}
	\graphicspath{{./figures/}}
\else
	\usepackage[dvips]{graphicx}
	\graphicspath{./figures/}
\fi
\usepackage{color}
\usepackage{pgf, tikz, pgfplots}
\usetikzlibrary{shapes, arrows, automata}
\usepackage{caption} 
\usepackage{subcaption} 
\usepackage{amsmath}
\usepackage{amsfonts, amssymb, amsthm}
\usepackage{mathrsfs}
\usepackage{upgreek}
\usepackage{algorithm,algpseudocode}
\usepackage{enumerate}
\usepackage{multirow}
\usepackage{rotating}
\usepackage{needspace}

\input{mysymbol.sty}

\newcommand{\QED}{\hfill\ensuremath{\blacksquare}}

\def\nv{{\mathrm{nv}}}
\def\ni{{\mathrm{ni}}}
\def\hv{{\mathrm{hv}}}

\title{CONVOLUTIONAL NEURAL NETWORKS VIA NODE-VARYING GRAPH FILTERS}
\name{Fernando Gama$^{\dag}$, Geert Leus$^{\ddag}$, Antonio G. Marques$^{\ast}$ and Alejandro Ribeiro$^{\dag}$\thanks{Supported by USA NSF CCF 1717120 and ARO W911NF1710438, and Spanish MINECO TEC2013-41604-R and TEC2016-75361-R. 
		}
}
\address{\dag \ Department of Electrical and Systems Engineering, University of Pennsylvania, Philadelphia, USA \\
\ddag \ Department of Microelectronics, Delft University of Technology, Delft, The Netherlands \\
$\ast$ \ Department of Signal Theory and Communications, King Juan Carlos University, Madrid, Spain
}
\begin{document}
\ninept
\maketitle
\begin{abstract}
Convolutional neural networks (CNNs) are being applied to an increasing number of problems and fields due to their superior performance in classification and regression tasks. Since two of the key operations that CNNs implement are convolution and pooling, this type of networks is implicitly designed to act on data described by regular structures such as images. Motivated by the recent interest in processing signals defined in irregular domains, we advocate a CNN architecture that operates on signals supported on graphs. The proposed design replaces the classical convolution not with a node-invariant graph filter (GF), which is the natural generalization of convolution to graph domains, but with a \textit{node-varying} GF. This filter extracts different local features without increasing the output dimension of each layer and, as a result, bypasses the need for a pooling stage while involving only local operations. A second contribution is to replace the node-varying GF with a \textit{hybrid node-varying} GF, which is a new type of GF introduced in this paper. While the alternative architecture can still be run locally  without requiring a pooling stage, the number of trainable parameters is smaller and can be rendered independent of the data dimension. Tests are run on a synthetic source localization problem and on the \texttt{20NEWS} dataset.
\end{abstract}
%

\section{Introduction} \label{sec_intro}
\vspace{-.15cm}
Convolutional neural networks (CNNs) have shown remarkable performance in a wide array of inference and reconstruction tasks \cite{lecun15-deeplearning}, in fields as diverse as pattern recognition, computer vision and medicine \cite{bruna13-scattering,lecun10-vision,greenspan16-medical}. The objective of CNNs is to find a computationally feasible architecture capable of reproducing the behavior of a certain unknown function. Typically, CNNs consist of a succession of layers, each of which performs three simple operations --$\,$usually on the output of the previous layer$\,$-- and feed the result into the next layer. These three operations are: 1) convolution, 2) application of a nonlinearity, and 3) pooling or downsampling. Because the classical convolution and downsampling operations are defined for regular (grid-based) domains, CNNs have been applied to act on data modeled by such a regular structure, like time or images.


However, an accurate description of modern datasets such as those in social networks or genetics \cite{lazer09-compsoc,davidson02-genetics} calls for more general irregular structures. A framework that has been gaining traction to tackle these problems is that of graph signal processing (GSP) \cite{sandryhaila13-dspg,sandryhaila14-freq,shuman13-mag}. GSP postulates that data can be modeled as a collection of values associated with the nodes of a graph, whose edges describe pairwise relationships between the data. By exploiting the interplay between the data and the graph, traditional signal processing concepts such as the Fourier transform, sampling and filtering have been generalized under the GSP framework to operate on a broader array of datasets \cite{chen15-selection,marques16-aggregation,segarra16-percolation}.

Motivated by the success of CNNs and the need to deal with irregular domains, recent efforts have been made to extend CNNs to work with data (signals) defined on manifolds and graphs \cite{bronstein16-geomdeeplearn}. Since in the GSP literature the notion of convolution is generalized to that of node-invariant graph filters (GFs) --matrix polynomials of the graph Laplacian--, existing CNN works operating on graph signals have replaced classical convolutions with such node-invariant GFs \cite{bruna14-deepspectralnetworks}. Nonetheless, how to generalize pooling remains elusive. Attempts using hierarchical multilayer clustering algorithms have been made \cite{defferrard17-cnngraphs}, but clustering is usually a computationally intensive operation \cite{carlsson10-hierarchical}.

This paper proposes a new architecture for CNNs operating on graph signals upon replacing convolutions with \textit{node-varying} GFs, which are more flexible local graph-signal operators described in \cite{segarra17-linear}. This not only introduces additional degrees of freedom, but also avoids the pooling stage and, as a result, the need to compute a cluster for each of the layers disappears. A second architecture is also proposed, that replaces convolutions with a \textit{hybrid node-varying} GF, a new graph-signal operator introduced in this paper that can be viewed as an intermediate design between node-varying and classical GFs. Our node-varying GF based architectures are able to extract different local features at varying resolutions, do not increase the dimension of the output of each layer, and can be implemented using only local exchanges.

\vspace{.1cm}
\noindent \textit{Paper outline:} Sec.~\ref{sec_cnn} reviews traditional CNNs and GSP and introduces the definition of node-varying and node-invariant GFs. Sec.~\ref{sec_localcnn} presents the new local graph CNN architectures using node-varying GFs. Sec.~\ref{sec_sims} runs tests on a synthetic source localization problem and on the \texttt{20NEWS} dataset. 

\section{Preliminaries: CNN and GSP} \label{sec_cnn}
\vspace{-.15cm}
Let $\bbx \in \ccalX$ be the input data or signal, defined on a field $\ccalX$, and let $\bby \in \ccalY$ be the output data, defined on a field $\ccalY$. Let $f: \ccalX \to \ccalY$ be a function such that $\bby = f(\bbx)$. Generically, the objective of CNNs is to design a function $\hhatf: \ccalX \to \ccalY$ such that a problem-dependent loss function $\ccalL(\bby,\hhatf(\bbx))$ is minimized. Standard choices for such a loss are the cross-entropy (for classification) or the mean square error (for regression). The function $\hhatf$ is built from a concatenation of $L$ layers $\hhatf = f_{L}\circ \cdots \circ f_{2} \circ f_{1}$ where each layer is a function $f_{\ell}: \ccalX_{\ell-1} \to \ccalX_{\ell}$, $\ell=1,\ldots,L$ with $\ccalX_{0}=\ccalX$ and $\ccalX_{L}=\ccalY$.   Each one of these layers is computed from three basic operations $\bbx_{\ell} = f_{\ell}(\bbx_{\ell-1}) = \ccalP_{\ell} \{ \rho_{\ell}(\ccalA_{\ell} (\bbx_{\ell-1})) \}$, where $\ccalA_{\ell}:\ccalX_{\ell-1} \to \ccalX_{\ell}'$ is a linear function, $\rho_{\ell}: \ccalX_{\ell}' \to \ccalX_{\ell}'$ is a nonlinear function, and $\ccalP_{\ell}: \ccalX_{\ell}' \to \ccalX_{\ell}$ is the pooling operator, and where $\bbx_{0}=\bbx$ and $\bbx_{L} = \hby = \hhatf(\bbx)$ is the estimated output after $L$ layers. It is noted that this architecture is computationally straightforward since it is comprised of simple operations, and it is also amenable to be efficiently trained by means of a back-propagation algorithm \cite{rumelhart86-backprop}.

In a CNN, the first operation of each layer is a \textit{convolution} with a filter $\ccalA_{\ell}( \bbx_{\ell-1}) = \bba_{\ell} \ast \bbx_{\ell-1}$. Filter $\bba_{\ell}$ has small support so that it acts as a computationally efficient local feature extractor by relating only a few nearby values of the signal. In order to extract several different features within the same region, a collection of $F_{\ell}$ filters $\{\bba_{\ell,k}\}_{k=1}^{F_{\ell}}$ is used, resulting in a $F_{\ell}$-times increase in the dimension of the output. To illustrate this with an example, consider that $\bbx=\bbx_{0}$ is an image of size $16 \times 16$, $\ccalX=\ccalX_{0} = \reals^{16 \times 16}$ and that, in the first layer, $\{\bba_{1,1},\ldots,\bba_{1,4}\}$ is a collection of $F_{1}=4$ filters of support $2 \times 2$ pixels. Then, $\ccalX_{1}' = \reals^{16 \times 16 \times 4}$ and $\ccalA_{1}: \reals^{16 \times 16} \to \reals^{16 \times 16 \times 4}$ with $\ccalA_{1} (\bbx) = \{\bba_{1,1} \ast \bbx, \ldots, \bba_{1,4} \ast \bbx\}$. 

The second operation is to apply a (pointwise) nonlinear function $\rho_{\ell}(\cdot)$ to the output of the linear step to yield $\rho_{\ell}(\ccalA_{\ell}(\bbx_{\ell-1}))\in \ccalX_{\ell}'$. The objective behind applying these nonlinearities at each layer is to create a structure flexible enough to reproduce general nonlinear behaviors. Typical choices for $\rho_{\ell}$ include rectified linear units (ReLUs) $\max\{0,x\}$ and the absolute value $|x|$ \cite{wiatowski17-maththeory}. Continuing with the previous example, now that the output of the convolution layer is $\ccalA_{1}(\bbx_{0}) \in \reals^{16 \times 16 \times 4}$, we apply a ReLU so that $\rho_{\ell}: \reals^{16 \times 16 \times 4} \to \reals^{16 \times 16 \times 4}$ with $[\rho_{\ell}(\ccalA_{1}\bbx_{0})]_{i,j,k} = \max(0,[\ccalA_{1}(\bbx_{0})]_{i,j,k})$ for $i,j=1,\ldots,16$ and $k=1,\ldots,4$.

The third operation is pooling, whose objective is twofold; i) given that each convolution operation increases the number of features, pooling keeps the output dimension under control; and ii) since it is desirable to analyze the data at different resolution levels, pooling reduces the distance between datapoints that were originally far away (with the reduction being more significant as more layers are added). It is noted that a better way to aggregate data in non-bandlimited signals is to do max-pooling or average-pooling instead of traditional downsampling \cite{wiatowski17-maththeory}. Returning to the ongoing example, assume that we consider max-pooling of size $2$. Then, $\ccalP_{1}: \reals^{16 \times 16 \times 4} \to \reals^{8 \times 8 \times 4}$ so that $\ccalX_{1}'=\reals^{16 \times 16 \times 4}$ and $\ccalX_{1} = \reals^{8 \times 8 \times 4}$ and where each element of $\bbx_{1}$ is obtained from computing the maximum value of $\rho_{1}(\ccalA_{1}(\bbx_{0}))$ within pixel masks of size $2\times 2$.

As already explained, those three operations are subsequently repeated by concatenating layers. The idea is to change the representation of the data by progressively trading samples for features \cite{jacobsen17-hierarchicalcn}. The target representation should be more useful for the specific task at hand as measured by the loss function $\ccalL$. The last step is typically a \emph{readout} layer implementing a (linear) map from $\ccalX_{L-1}$ to $\ccalY$.

\vspace{.1cm}
\noindent \textit{Remark:} Albeit fairly typical, modifications to the described CNN architecture have been developed. These range from using outputs of different layers as input to the next layer \cite{huang17-densecnn}, to assuming that the useful output is collected at every layer instead of the last one \cite{bruna13-scattering}, to adding fully-connected layers after reaching the all-feature vector \cite{kuo17-recos}. Also, avoiding the pooling stage has been discussed \cite{springenberg15-nopooling}.



\subsection{Graph signals and filters} \label{subsec_gsp}
\vspace{-.1cm}
In this paper, we consider each datapoint in the dataset to be modeled as a graph signal. To be specific, let $\ccalG=(\ccalV,\ccalE,\ccalW)$ be a graph with a node set $\ccalV$ with cardinality $N$, a set of edges $\ccalE \subseteq \ccalV \times \ccalV$, and a weight function $\ccalW: \ccalE \to \reals$. A graph signal is then a mapping $\bbx: \ccalV \to \reals$ that assigns a real number to each node and can be  conveniently represented as a vector $\bbx \in \reals^N$, with element $[\bbx]_{k}$ being the signal value at node $k$. Modeling a dataset as a graph signal allows for arbitrary pairwise relationships between the elements of the datapoint (i.e. between the elements of the vector). This relationship is brought to the fore by means of a graph shift operator (GSO) $\bbS \in \reals^{N \times N}$ which is the matrix that relates the signal with the underlying graph support. More specifically, $\bbS$ is such that $[\bbS]_{ij} \neq 0$ only if $(i,j) \in \ccalE$ or if $i=j$. This means that $\bbS \bbx$ is a local computation that can be carried out by operating only on the neighborhood. Examples of GSOs are the adjacency matrix, the graph Laplacian and their normalized counterparts \cite{sandryhaila14-freq,shuman13-mag}.

The GSO is the key to define the graph Fourier transform (GFT) and the different types of GFs. Assuming first that $\bbS = \bbV \bbLambda \bbV^{H}$ is a normal matrix diagonalized by a unitary matrix $\bbV$, the GFT of a signal $\bbx$ is defined as $\tbx = \bbV^{H}\bbx$. Moreover, node-invariant and node-varying GFs are defined, respectively, as \cite{segarra17-linear}
\begin{equation} \label{eqn_gf}
	\textstyle \bbH_{\ni} := \sum_{t=0}^{T-1} h_{t} \bbS^{t}, \qquad \qquad \bbH_{\nv} := \sum_{t=0}^{T-1} \diag(\bbh_{t}) \bbS^{t},
\end{equation}
where $T$ is the order of the filter, and $\{h_t\}_{t=0}^{T-1}$ and $\{\bbh_{t}\}_{t=0}^{T-1}$ are the filter coefficients. Furthermore, if $\bbh_{t} \in \reals^{N} $ is set such that $[\bbh_{t}]_k=h_t$ for all $k$ and $t$, then filter $\bbH_{\nv}$ reduces to $\bbH_{\ni}$.

Two interesting properties of the GFs in \eqref{eqn_gf} are: i) they are linear operators that account for the structure of the graph via $\bbS$, and ii) since $\bbS$ is a local (one-hop) operator and the output of either $\bbH_{\nv}$ or $\bbH_{\ni}$ can be viewed as a linear combination of successive applications of $\bbS$ to the input, it follows that $\bbH_{\nv}$ or $\bbH_{\ni}$ are local operators as well. The main difference is that while $\bbS$ takes into account information within the one-hop neighborhood of the nodes, the operators in \eqref{eqn_gf} consider information that is within their $T\!-\!1$ neighborhood \cite{segarra17-linear}. 



\subsection{CNNs using node-invariant GFs} \label{subsec_related}

Recent efforts have been made towards extending CNNs to operate on graph signals in the hope of carrying over their excellent performance to a broader class of problems (see \cite{bronstein16-geomdeeplearn} for a general survey). The existing works typically set the GSO as the graph Laplacian matrix and, more importantly, replace the classical convolutions with node-invariant GFs [cf. $\bbH_{\ni}$ in \eqref{eqn_gf}]. The main reason for this is that node-invariant GFs allow for the generalization of the convolution theorem to graph signals in the sense that filtering in the (node) domain implies multiplication in the frequency domain given by the GFT. To see why this is the case, consider the graph signal $\bby = \bbH_{\ni} \bbx$, recall the eigendecomposition of the GSO $\bbS$, and note that since $\bbH_{\ni}$ is a matrix polynomial on $\bbS$, its eigenvectors are also $\bbV$. With these considerations, after applying the GFT to the input-output equation $\bby = \bbH_{\ni} \bbx$ we have that $\tby = \diag(\tbh) \tbx$ with $\diag(\tbh) := \sum_{t=0}^{T-1} h_{t} \bbLambda^{t}$ being the filter's frequency response.

Building on this interpretation, \cite{bruna14-deepspectralnetworks} designed the filter coefficients to be used at each layer in the spectral domain. To avoid the (expensive) computation of eigendecompositions, a Chebyshev approximation which operates in the node domain using a low-order node-invariant GF was adopted in \cite{defferrard17-cnngraphs}. While convolutions have been replaced with node-invariant GFs and  point-wise nonlinearities with node-wise nonlinearities applied locally at each node of the graph, there is no consensus on how pooling must be implemented. 
The suggestion in \cite{bruna14-deepspectralnetworks} was to use multiscale hierarchical algorithms to create a collection of related graphs with less and less nodes. In that context, \cite{defferrard17-cnngraphs} adopted the Graclus algorithm \cite{dhillon07-graclus} and suggested an innovative pooling system by means of a binary tree partition. It is noted that clustering is in itself an ill-posed problem and that there exist several criteria for determining \emph{good} clusters \cite{horta15-compareclusters,gama17-cutmetrics}. Moreover, it is usually a computationally intensive operation \cite{carlsson10-hierarchical,ward63-linkage,defays76-complete}.


\section{CNN Architecture using Node-Varying GF} \label{sec_localcnn}

Starting from the CNN architecture described in Sec. \ref{sec_cnn}, we propose a new architecture for CNNs that at each layer $\ell$: 1) replaces convolutions with node-varying GFs [cf.$\,\bbH_{\nv}\,$in$\,$\eqref{eqn_gf}]; 2) applies a local node-wise nonlinearity; and 3) does not apply a pooling stage, thus avoiding the computation of clusters for each of the layers. 

To motivate the proposed design, recall that the idea in the convolution stage is to get several features per region and, for that, $F_{\ell}$ filters are employed. This naturally increases the dimension of the signal by a factor of $F_{\ell}$ and pooling becomes necessary to prevent a geometric growth of the size of the data. That is, there is a trade-off between the availability of multi-resolution features extracted from the data and the size of the information passed onto the next layer. Our proposed architecture tries to extract local features at different locations of the graph without increasing dimensionality. Being more specific, by adopting the node-varying GF in \eqref{eqn_gf}, each node gains the ability to weight their local neighborhood differently, and because nodes within a neighborhood weight differently their respective neighborhoods, each of them acts as a different feature within the region. Since the output of a node-varying GF is another graph signal, then the dimensionality of the data at each layer is not increased while local features are captured respectively by each node. The data analysis at different resolutions comes naturally with the adoption of this kind of filters and is adjusted by the length of the filters on each layer. Concretely, by applying a filter of length $T_{1}$ each node gathers information of up to the $T_{1}-1$ neighborhood; then, in the following layer, when another filter of length $T_{2}$ is applied, then nodes actually disseminate information up to the $T_{2}-1$ neighborhood from the previous layer, so that the total information processed goes up to the $T_{1}+T_{2}-2$ neighborhood. Therefore, as the local graph CNN goes deeper, it gathers more global information. 

\vspace{-0.3cm}

\subsection{CNN via hybrid node-varying GFs}
A key aspect of any CNN architecture is the number of parameters that need to be optimized in the training phase \cite{huang17-densecnn}. Based on this criterion, it is observed that adopting a node-varying GF results in a number of parameters proportional to the number of nodes, the length of the filter at each layer and the number of layers $\sum_{l=1}^LNT_l$. This might be an undesirable characteristic of the architecture, especially for high-dimensional datasets. In order to overcome this, we propose an alternative design where the convolution is replaced with a \emph{hybrid node-varying GF}. 

To define this new type of GF, start by considering a \textit{tall} binary matrix $\bbC_{\ccalB} \in \{0,1\}^{N \times B}$ with exactly one non-zero entry per row. Define now the reduced vector of filter coefficients as $\bbh_{\ccalB,t} \in \reals^B$. Then a hybrid node-varying GF is a graph signal operator of the form
\begin{equation} \label{eqn_filter_taps}
\textstyle \bbH_{\hv} := \sum_{t=0}^{T-1} \diag(\bbC_{\ccalB} \bbh_{\ccalB,t}) \bbS^{t}. 
\end{equation}
Clearly the GF above is linear, accounts for the structure of the graph, and can be implemented locally. The name ``hybrid'' is due to the fact that i) if $B=N$ and $\bbC_{\ccalB}=\bbI$, then $\bbH_{\hv}$ is equivalent to $\bbH_{\nv}$; and ii) if $B=1$, then $\bbH_{\hv}$ reduces to $\bbH_{\ni}$. 

While basis expansion models other than $\bbh_t=\bbC_{\ccalB}\bbh_{\ccalB,t}$ could have been used, $\bbC_{\ccalB}$ was selected to be binary to facilitate intuition and keep implementation simple.  In particular, the columns of $\bbC_{\ccalB}$ can be viewed as membership indicators that map nodes into different groups. With this interpretation, $\{[\bbh_{\ccalB,t}]_b\}_{t=0}^{T-1}$ represents the common filter coefficients that each node of the $b$th group will use. This demonstrates that the selection of the method to group the nodes offers a new degree of freedom for the design of \eqref{eqn_filter_taps} and the corresponding CNN. Different from the multi-resolution clustering algorithms associated with the pooling stage, this algorithm performs a single grouping. In the simulations presented in the next section, the grouping implicit in $\bbC_{\ccalB}$ is carried out in two steps. First, we form the set  $\ccalB = \{v_{1},\ldots,v_{B}\}$ containing the $B$ nodes with the highest degree (ties are broken uniformly at random) and set $[\bbC_{\ccalB}]_{v_{b},b}=1$ for all $b=1,...,B$. Second, for all the nodes that do not belong to $\ccalB$ we set the membership matrix as 
 \begin{equation} \label{eqn_copy_matrix}
 \textstyle [\bbC_{\ccalB}]_{ij} = 1 
 \textrm{ if }
 j \in \argmax_{b : v_{b} \in \ccalB} \{\ccalW(i,v_{b})\} \ , \ i\notin \ccalB,
 \end{equation}
 where $\ccalW(i,v_{b})$ is the edge weight. That is, for each of the nodes not in $\ccalB$ we copy the filter coefficients of the node in $\ccalB$ that exercises the largest influence. As before, ties are broken uniformly at random.
CNN schemes with region-dependent filters have been used in the context of images using regular convolutions \cite{gregor10-localreceptivefields, huang12-hierarchicalface}. The regional features computed at each layer are kept separate and only the last stages (involving fully connected layers) merge them. The use of node-varying graph filters proposed in this paper, not only changes the definition of the convolution, but also merges the regional features at every layer.
 
\vspace{.1cm}
\noindent \textit{CNN architecture:} Adopting the hybrid node-varying GF for the first stage of each layer of our CNN implies that the total number of parameters to be learned is $\sum_{l=1}^L B T_{l}$, which is independent of $N$ and guarantees that the proposed architecture scales well for high-dimensional data. Lower values of $B$ will decrease the number of training parameters, while limiting the ability of extracting features of the filter.  All in all, the architecture of the proposed CNN is given by Algorithm~\ref{algm_local-gcnn}. We observe that, except for the final readout layer, all computations are carried out in a local fashion making the CNN amenable to a distributed implementation. 
%
%
%
Finally, let us note that while some problems inherently live in a constant-dimension submanifold and make the choice of a constant $B$ possible, some other problems might have a lower dimension that still grows with $N$ but in a sublinear fashion. Therefore, while $B$ might not be independent of $N$, it could still be chosen as a sublinear function of $N$ \cite{wakin05,verma12}.

\begin{algorithm}[t]
 	\caption{(Hybrid) Node-varying GF CNN.}
	\label{algm_local-gcnn}
	
	\begin{algorithmic}[1]
 \Statex \textbf{Input:} $\{\bbx\}$: test dataset, $\{(\bbx',\bby')\}$: train dataset
 \Statex $\quad \bbS$: GSO, $\{T_{1},\ldots,T_{L-1}\}$: degrees of layer
 \Statex $\quad B$: number of nodes to select for weights
 \Statex \textbf{Output:} $\{\hby\}$: estimates
 \Statex
 \Procedure{nvgf\_cnn}{$\{\bbx\}$,$\{(\bbx',\bby')\}$,$\bbS$,$\{T_{1},\ldots,T_{L-1}\}$,$B$}
   \State Create set $\ccalB$ by selecting $B$ nodes with highest degree
   \State Compute $\bbC_{\ccalB}$ \Comment{See \eqref{eqn_copy_matrix}}
   \State Create the $L-1$ layers:
   \For{$\ell = 1:L-1$}
       \State Create $B$ filter taps $\{\bbh_{\ccalB,0},\ldots,\bbh_{\ccalB,T_{\ell}-1}\}$
       \State Obtain $\bbH_{\ell} = \sum_{t=0}^{T_{\ell}-1} \diag(\bbC_{\ccalB} \bbh_{\ccalB,t}) \bbS^{t}$ \Comment{See \eqref{eqn_filter_taps}}
       \State Apply non-linearity $\rho_{\ell}(\bbH_{\ell} \ \cdot)$
   \EndFor
   \State Create readout layer $\ccalA_{L} \ \cdot$
   \State Learn $\{\bbh_{\ccalB,0},\ldots,\bbh_{\ccalB,T_{\ell-1}}\}_{\ell=1}^{L-1}$ and $\ccalA_{L}$ from $\{(\bbx',\bby')\}$
   \State Obtain $\hby = \hbf(\bbx)$ using trained coefficients
 \EndProcedure
 	\end{algorithmic}
\end{algorithm}

\begin{figure*}
\centering
\begin{subfigure}{.33\textwidth}
  \centering
  \vspace{-0.15in}
  \includegraphics[width=0.9\textwidth]{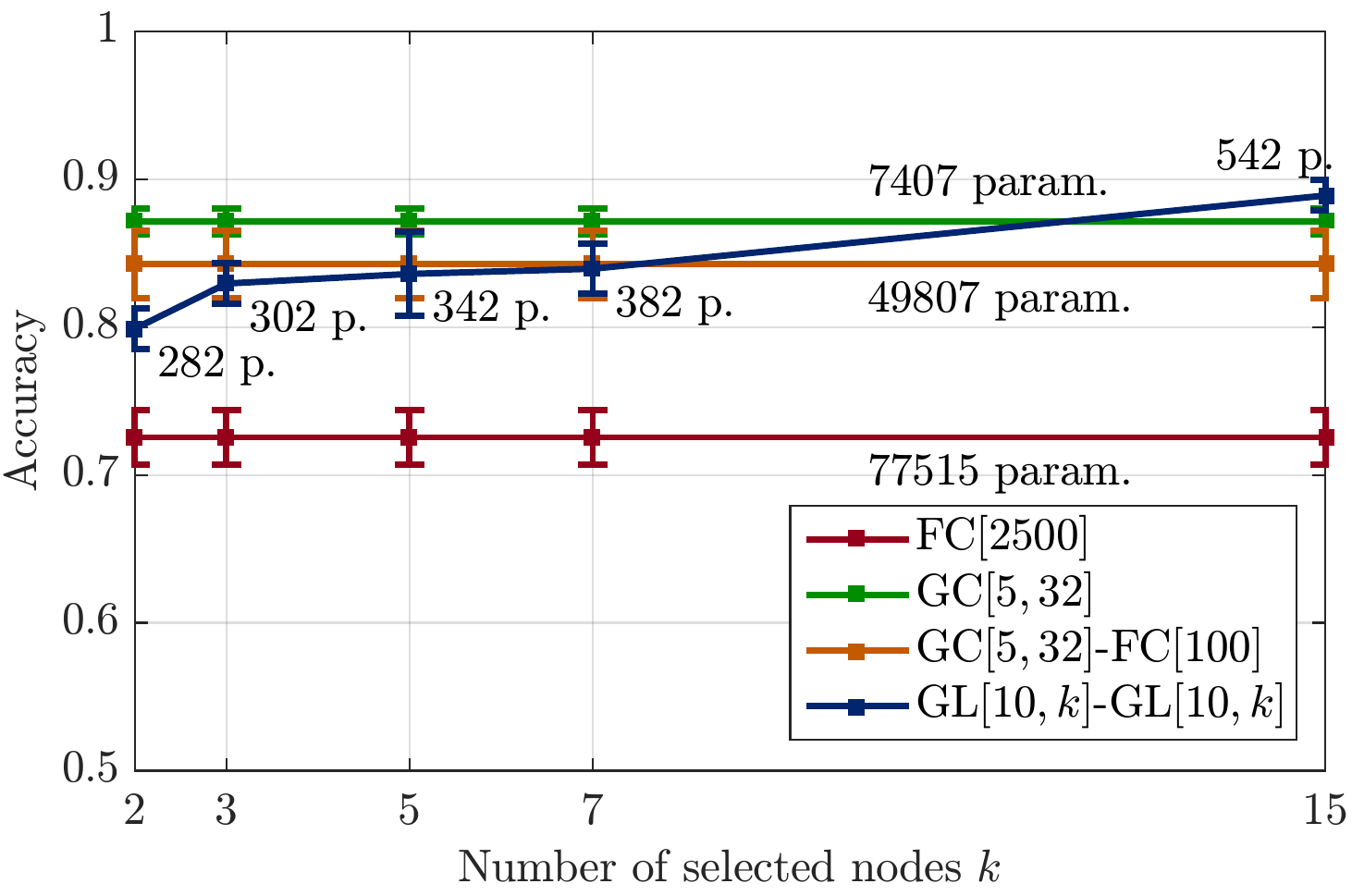}
   \vspace{-0.05in}
  \caption{}
  \label{fig_varF}
\end{subfigure}%
\begin{subfigure}{.33\textwidth}
  \centering
    \vspace{-0.15in}
  \includegraphics[width=0.9\textwidth]{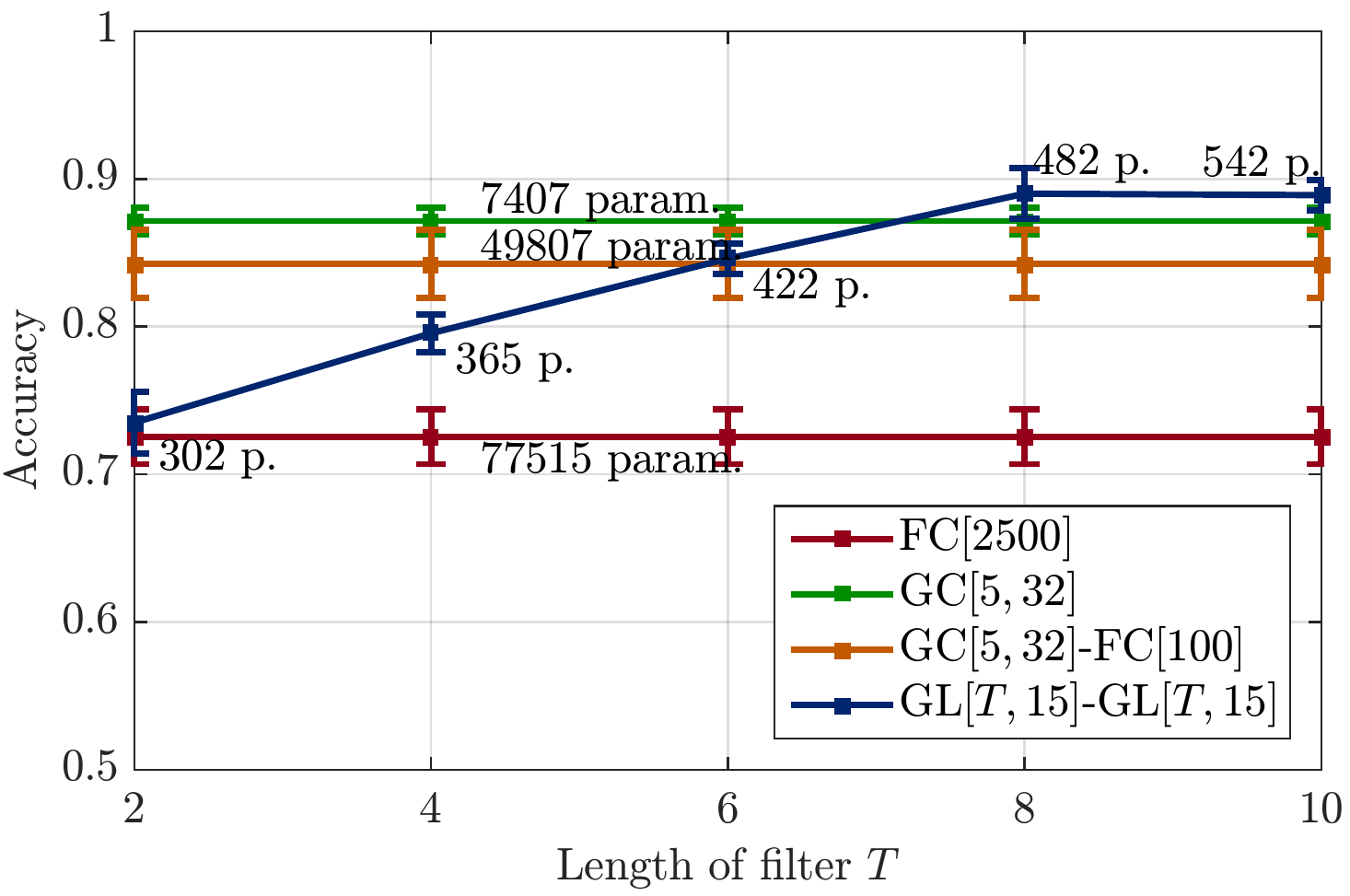}
   \vspace{-0.05in}
  \caption{}
  \label{fig_varK}
\end{subfigure}%
\begin{subfigure}{.33\textwidth}
  \centering
    \vspace{-0.15in}
  \includegraphics[width=0.9\textwidth]{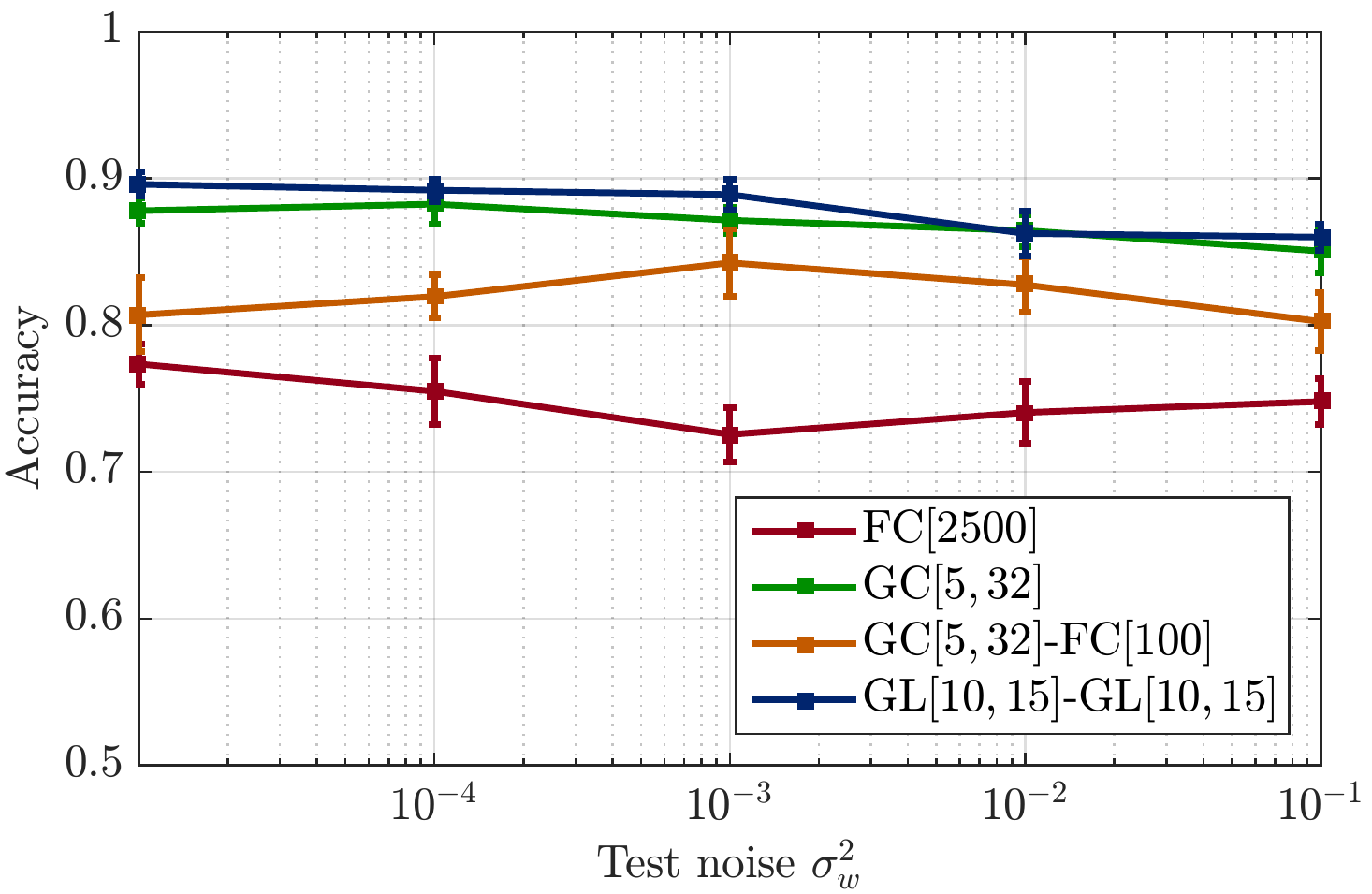}
  \vspace{-0.05in}
  \caption{}
  \label{fig_varsigma2}
\end{subfigure}%

\vspace{-0.12in}

\caption{{\footnotesize Accuracy in the source localization problem. Results were averaged across $10$ different realizations. For clarity of figures, error bars represent $1/4$ of the estimated variance. The number of parameters of each architecture is also shown. (\subref{fig_varF}) As a function of the number of selected nodes $k$. Accuracy gets better as more and more nodes are selected to extract features. (\subref{fig_varK}) As a function of the length of the filter $T$. Accuracy improves when filters are longer. It is observed that after length $8$ for which $2$ layers are able to obtain all the relevant information in the graph ($15$ nodes), accuracy does not improve substantially. (\subref{fig_varsigma2}) As a function of the noise in the test set. The proposed architecture is fairly robust to noise since accuracy drops approximately $5\%$ across $5$ orders of magnitude.}}
\vspace{-0.05in}
\label{fig_sourceloc}
\end{figure*}

\section{Numerical tests} \label{sec_sims}

In this section, we run tests on the proposed CNN architecture and compare it with the one developed in \cite{defferrard17-cnngraphs}. In general, we observe that our CNN achieves similar performance but with at least one order of magnitude less of parameters. In the first testcase we consider a synthetic dataset of a source localization problem in which different diffused graph signals are processed to determine the single node that originated them. In the second testcase we use the \texttt{20NEWS} dataset and a \texttt{word2vec} embedding underlying graph to classify articles in one out of $20$ different categories \cite{joachims96-20news}. For both problems, denote as GC$[T,k]$ a graph CNN using Chebyshev polynomial approximation of order $T$ with $k$ features; as FC$[k]$ a fully connected layer CNN with $k$ hidden units; and as GL$[T,k]$ the proposed CNN where the degree-based hybrid node-varying GF is of order $T$ with $B=k$ nodes selected. A ReLU nonlinearity is applied at each layer and all architectures include a readout layer. We note that the total parameter count includes this last readout layer as well as bias parameters typically used before applying the ReLU nonlinearity. For the training stage in both problems, an ADAM optimizer with learning rate $0.005$ was employed \cite{kingma17-adam}, for $20$ epochs and batch size of $100$.


\vspace{0.075cm}
\noindent  \textbf{Testcase 1: Source localization.} 
Consider a connected Erd\H{o}s-R\'{e}nyi (ER) graph with $N$ nodes and edge probability $p_{ER}=0.4$ and let $\bbW$ denote its adjacency matrix. With $\bbdelta_{c}$ representing a graph signal taking the value $1$ at node $c$ and $0$ elsewhere, the signal $\bbx = \bbW^{t} \bbdelta_{c}$ is a diffused version of the sparse input $\bbdelta_{c}$ for some unknown $0 \leq t \leq N-1$. The objective is to determine the source $c$ that originated the signal $\bbx$ irrespective of time $t$. To that end, we create a set of $N_{\textrm{train}}$ labeled training samples $\{(c',\bbx')\}$ where $\bbx' = \bbW^{t} \bbdelta_{c'}$ with both $c'$ and $t$ chosen at random. Then we create a test set with $N_{\textrm{test}}$ samples in the same fashion, but we add i.i.d. zero-mean Gaussian noise $\bbw$ with variance $\sigma_{w}^{2}$, so that the signals to be classified are $ \bbW^{t} \bbdelta_{c} + \bbw$. The goal is to use the training samples to design a CNN that determines the source (node) $c$ that originated the diffused.

For a graph with $N=15$ nodes we test four architectures: (a) FC$[2500]$, (b) GC$[5,32]$, (c) GC$[5,32]$-FC$[100]$ and (d) GL$[10,15]$-GL$[10,15]$. The GSO employed is the adjacency matrix $\bbS=\bbW$. A dropout of $0.5$ is included in the training phase. The test set is of size $N_{\textrm{test}}=200$. Results are listed in Table~\ref{table_sourceloc}. Note that these results are obtained by averaging $10$ different realizations of the problem. We observe that the performance of our CNN is similar to that of GC$[5,32]$ but with ten times less parameters.

\begin{table}
	\centering
\begin{tabular}{lrc} \hline
Architecture 			& Parameters 	& Accuracy 	\\ \hline
FC$[2500]$				& $77,515$		& $72.6\%$	\\ 
GC$[5,32]$				& $ 7,407$		& $87.2\%$	\\
GC$[5,32]$-FC$[100]$	& $49,807$		& $84.3\%$	\\
GL$[10,15]$-GL$[10,15]$	& $   542$		& $88.9\%$	\\ \hline
\end{tabular}
 \vspace{-0.1in}
	\caption{{\footnotesize Source localization results for $N=15$ nodes.}}
	\vspace{-0.2in}
	\label{table_sourceloc}
\end{table}

Additionally, we run tests changing the values of several of the parameters of the architecture. In Fig.~\ref{fig_varF} we observe the accuracy obtained when varying the number of selected nodes. It is noted that selecting less nodes implies that less features are extracted. This impacts negatively the accuracy. Nonetheless, even an accuracy level of $80\%$ is achieved with as few as $282$ parameters, which is a better performance than using a fully connected layer with $2500$ hidden units which requires $100$ times more parameters. The dependence of the accuracy on the length of the filter $T$ can be observed in Fig.~\ref{fig_varK}. We note a linear increase in accuracy that saturates around $T=8$. This is the length for which, when using two layers, the information corresponding to the whole graph can be aggregated. Finally, in Fig.~\ref{fig_varsigma2} we show the performance of all four architectures as a function of the noise on the test set. We observe that both GC$[5,32]$ and the proposed architecture achieve similar accuracies with a fairly robust performance, since the accuracy dropped only $5\%$ within a $5$ order magnitude change in the noise. Values shown are mean accuracies obtained after averaging $10$ realizations and the error bars represent $1/4$ of the estimated variance from these realizations.


\vspace{.075cm}
\noindent  \textbf{Testcase 2: \texttt{20NEWS} dataset.} 
\begin{table}
	\centering
\begin{tabular}{lrc} \hline
Architecture 			& Parameters 	& Accuracy 	\\ \hline
GC$[5,32]$				& $1,920,212$	& $60.75\%$	\\
GL$[5,1500]$			& $  67,521$	& $60.34\%$	\\ \hline
\end{tabular}
 \vspace{-0.1in}
	\caption{{\footnotesize Results for classification on \texttt{20NEWS} dataset on a \texttt{word2vec} graph embedding of $N=3,000$ nodes.}}
	\vspace{-0.2in}
	\label{table_20news}
\end{table}
Here we consider the classification of articles in the \texttt{20NEWS} dataset which consists of $18,846$ texts ($11,314$ of which are used for training and $7,532$ for testing) \cite{joachims96-20news}. The graph signals are constructed as in \cite{defferrard17-cnngraphs}: each document $x$ is represented using a normalized bag-of-words model and the underlying graph support is constructed using a $16$-NN graph on the \texttt{word2vec} embedding \cite{mikolov13-word2vec} considering the $3,000$ most common words. The GSO adopted is the normalized Laplacian. No dropout is used in the training phase. The architectures used are GC$[5,32]$ and GL$[5,1500]$. Accuracy results are listed in Table~\ref{table_20news}, demonstrating that both architectures achieve similar accuracies, but with our CNN requiring $100$ times less parameters.

\vspace{-0.3cm}

\section{Conclusions} \label{sec_conclusions}

\vspace{-0.25cm}

A CNN architecture to operate on graph signals was proposed. The convolution stage was replaced by a node-varying GF, and no pooling stage was implemented. Extraction of different features was achieved by the adoption of a node-varying GF and resolution levels were adjusted via the length of the filter. The convolutional layers of the resulting CNN could be implemented locally. To prevent the number of parameters to grow with the size of the data, we proposed a hybrid node-varying GF where nodes were grouped and the same filter coefficients were used within a particular group. Results on the \texttt{20NEWS} dataset showed a performance similar to that of existing CNNs implementing node-invariant GFs but with $100$ times less parameters to train. A synthetic source localization problem was used to asses numerically the sensitivity of the estimation performance with respect to the number of groups and the degree of the filter.


\bibliographystyle{IEEEbib}
\bibliography{myIEEEabrv,bib-pooling}

\begin{thebibliography}{10}

\bibitem{lecun15-deeplearning}
Y.~LeCun, Y.~Bengio, and G.~Hinton,
\newblock ``Deep learning,''
\newblock {\em Nature}, vol. 521, no. 7553, pp. 85--117, 2015.

\bibitem{bruna13-scattering}
J.~Bruna and S.~Mallat,
\newblock ``Invariant scattering convolution networks,''
\newblock {\em {IEEE} Trans. Pattern Anal. Mach. Intell.}, vol. 35, no. 8, pp.
  1872--1886, Aug. 2013.

\bibitem{lecun10-vision}
Y.~LeCun, K.~Kavukcuoglu, and C.~Farabet,
\newblock ``Convolutional networks and applications in vision,''
\newblock in {\em 2010 {IEEE} Int. Symp. Circuits and Syst.}, Paris, France, 30
  May-2 June 2010, IEEE.

\bibitem{greenspan16-medical}
H.~Greenspan, B.~van Ginneken, and R.~M. Summers,
\newblock ``Deep learning in medical imaging: Overview and future promise of an
  exciting new technique,''
\newblock {\em {IEEE} Trans. Med. Imag.}, vol. 35, no. 5, pp. 1153--1159, May
  2016.

\bibitem{lazer09-compsoc}
{D. Lazer et al.},
\newblock ``Life in the network: The coming age of computational social
  science,''
\newblock {\em Science}, vol. 323, no. 5915, pp. 721--723, Feb. 2009.

\bibitem{davidson02-genetics}
E.~H.~Davidson et~al.,
\newblock ``A genomic regulatory network for development,''
\newblock {\em Science}, vol. 295, no. 5560, pp. 1669--1678, Feb. 2002.

\bibitem{sandryhaila13-dspg}
A.~Sandryhaila and J.~M.~F. Moura,
\newblock ``Discrete signal processing on graphs,''
\newblock {\em {IEEE} Trans. Signal Process.}, vol. 61, no. 7, pp. 1644--1656,
  Apr. 2013.

\bibitem{sandryhaila14-freq}
A.~Sandyhaila and J.~M.~F. Moura,
\newblock ``Discrete signal processing on graphs: Frequency analysis,''
\newblock {\em {IEEE} Trans. Signal Process.}, vol. 62, no. 12, pp. 3042--3054,
  June 2014.

\bibitem{shuman13-mag}
D.~I Shuman, S.~K. Narang, P.~Frossard, A.~Ortega, and P.~Vandergheynst,
\newblock ``The emerging field of signal processing on graphs: Extending
  high-dimensional data analysis to networks and other irregular domains,''
\newblock {\em {IEEE} Signal Process. Mag.}, vol. 30, no. 3, pp. 83--98, May
  2013.

\bibitem{chen15-selection}
S.~Chen, R.~Varma, A.~Sandryhaila, and J.~Kova{\v{c}}evi{\' c},
\newblock ``Discrete signal processing on graphs: Sampling theory,''
\newblock {\em {IEEE} Trans. Signal Process.}, vol. 63, no. 24, pp. 6510--6523,
  Dec. 2015.

\bibitem{marques16-aggregation}
A.~G.~Marques, S.~Segarra, G.~Leus, and A.~Ribeiro,
\newblock ``Sampling of graph signals with successive local aggregations,''
\newblock {\em {IEEE} Trans. Signal Process.}, vol. 64, no. 7, pp. 1832--1843,
  Apr. 2016.

\bibitem{segarra16-percolation}
S.~Segarra, A.~G.~Marques, G.~Leus, and A.~Ribeiro,
\newblock ``{Reconstruction of Graph Signals Through Percolation from Seeding
  Nodes},''
\newblock {\em {IEEE} Trans. Signal Process.}, vol. 64, no. 16, pp. 4363--4378,
  Aug. 2016.

\bibitem{bronstein16-geomdeeplearn}
M.~M. Bronstein, J.~Bruna, Y.~LeCun, A.~Szlam, and P.~Vandergheynst,
\newblock ``{Geometric Deep Learning: Going Beyond Euclidean Data},''
\newblock {\em arXiv:1611.08097v1 [cs.CV]}, 24 Nov. 2016.

\bibitem{bruna14-deepspectralnetworks}
J.~Bruna, W.~Zaremba, A.~Szlam, and Y.~LeCun,
\newblock ``Spectral networks and deep locally connected networks on graphs,''
\newblock {\em arXiv:1312.6203v3 [cs.LG]}, 21 May 2014.

\bibitem{defferrard17-cnngraphs}
M.~Defferrard, X.~Bresson, and P.~Vandergheynst,
\newblock ``Convolutional neural networks on graphs with fast localized
  spectral filtering,''
\newblock {\em arXiv:1606.09375v3 [cs.LG]}, 5 Feb. 2017.

\bibitem{carlsson10-hierarchical}
G.~Carlsson and F.~M{\'{e}}moli,
\newblock ``Characterization, stability and convergence of hierarhical
  clustering methods,''
\newblock {\em J. Mach. Learning Res.}, vol. 11, pp. 1425--1470, Apr. 2010.

\bibitem{segarra17-linear}
S.~Segarra, A.~G.~Marques, and A.~Ribeiro,
\newblock ``Optimal graph-filter design and applications to distributed linear
  network operators,''
\newblock {\em {IEEE} Trans. Signal Process.}, vol. 65, no. 15, pp. 4117--4131,
  Aug. 2017.

\bibitem{rumelhart86-backprop}
D.~E. Rumelhart, G.~E. Hinton, and R.~J. Williams,
\newblock ``Learning representations by back-propagating errors,''
\newblock {\em Nature}, vol. 323, no. 6088, pp. 533--536, Oct. 1986.

\bibitem{wiatowski17-maththeory}
T.~Wiatowski and H.~B{\"{o}l}cskei,
\newblock ``A mathematical theory of deep convolutional neural networks for
  feature extraction,'' Website, 23 March 2017.

\bibitem{jacobsen17-hierarchicalcn}
J.-H. Jacobsen, E.~Oyallon, S.~Mallat, and A.~W.~M. Smeulders,
\newblock ``Multiscale hierarchical convolutional networks,''
\newblock {\em arXiv:1703.04140v1 [cs.LG]}, 12 March 2017.

\bibitem{huang17-densecnn}
G.~Huang, Z.~Liu, L.~van~der Maaten, and K.~Q. Weinberger,
\newblock ``Densely connected convolutional networks,''
\newblock {\em arXiv:1608.06993v4 [cs.CV]}, 27 Aug. 2017.

\bibitem{kuo17-recos}
C.-C.~J. Kuo,
\newblock ``The cnn as a guided multilayer recos transform,''
\newblock {\em {IEEE} Signal Process. Mag.}, vol. 34, no. 3, pp. 81--89, May
  2017,
\newblock lecture notes.

\bibitem{springenberg15-nopooling}
J.~T. Springenberg, A.~Dosovitskiy, T.~Brox, and M.~Riedmiller,
\newblock ``Striving for simplicity: The all convolutional net,''
\newblock {\em arXiv:1412.6806v3 [cs.LG]}, 13 Apr. 2015.

\bibitem{dhillon07-graclus}
I.~Dhillon, Y.~Guan, and B.~Kulis,
\newblock ``Weighted graph cuts without eigenvectors: A multilevel approach,''
\newblock {\em {IEEE} Trans. Pattern Anal. Mach. Intell.}, vol. 29, no. 11, pp.
  1944--1957, Nov. 2007.

\bibitem{horta15-compareclusters}
D.~Horta and R.~J. G.~B. Campello,
\newblock ``Comparing hard and overlapping clusterings,''
\newblock {\em J. Mach. Learning Res.}, vol. 16, pp. 2949--2997, Dec. 2015.

\bibitem{gama17-cutmetrics}
F.~Gama, S.~Segarra, and A.~Ribeiro,
\newblock ``{Hierarchical Overlapping Clustering of Network Data Using Cut
  Metrics},''
\newblock {\em {IEEE} Trans. Signal, Inform. Process. Networks}, vol. PP, no.
  99, 24 May 2017.

\bibitem{ward63-linkage}
J.~H. Ward~Jr.,
\newblock ``Hierarchical grouping to optimize an objective function,''
\newblock {\em J. Amer. Statist. Assoc.}, vol. 58, no. 301, pp. 236--244, March
  1963.

\bibitem{defays76-complete}
D.~Defays,
\newblock ``An efficient algorithm for a complete link method,''
\newblock {\em Comput. J.}, vol. 20, no. 4, pp. 364--366, Apr. 1977.

\bibitem{gregor10-localreceptivefields}
K.~Gregor and Y.~LeCun,
\newblock ``Emergence of complex-like cells in a temporal product network with
  local receptive fields,''
\newblock {\em arXiv:1006.0448v1 [cs.NE]}, 2 June 2010.

\bibitem{huang12-hierarchicalface}
G.~B. Huang, H.~Lee, and E.~Learned-Miller,
\newblock ``Learning hierarchical representations for face verification with
  convolutional deep belief networks,''
\newblock in {\em 2012 {IEEE} Conf. Comput. Vision, Pattern Recognition},
  Providence, RI, 16-21 June 2012, IEEE Comput. Soc.

\bibitem{wakin05}
M.~B. Wakin, D.~L. Donoho, H.~Choi, and R.~G. Baraniuk,
\newblock ``The multiscale structure of non-differentiable image manifolds,''
\newblock in {\em SPIE Optics + Photonics}, San Diego, CA, July 2005, SPIE.

\bibitem{verma12}
N.~Verma, S.~Kpotufe, and S.~Dasgupta,
\newblock ``Which spatial partition trees are adaptive to intrinsic
  dimension?,''
\newblock {\em arXiv:1205.2609v1 [stat.ML]}, 9 May 2012.

\bibitem{joachims96-20news}
T.~Joachims,
\newblock ``Analysis of the rocchio algorithm with tfidf for text
  categorization,''
\newblock Computer Science Technical Report CMU-CS-96-118, Carnegie Mellon
  University, 1996.

\bibitem{kingma17-adam}
D.~P. Kingma and J.~L. Ba,
\newblock ``Adam: A method for stochastic optimization,''
\newblock {\em arxiv:1412.06980v9 [cs.LG]}, 30 Jan. 2017.

\bibitem{mikolov13-word2vec}
T.~Mikolov, K.~Chen, G.~Corrado, and J.~Dean,
\newblock ``Efficient estimation of word representations in vector space,''
\newblock {\em arXiv:1301.3781v3}, 7 Sep. 2013.

\end{thebibliography}

\end{document}